\documentclass{article}
\usepackage{spconf,amsmath,graphicx}
\usepackage[caption=false,font=footnotesize]{subfig}
\usepackage{multirow}

\usepackage{xspace}

\newcommand{\OneDot}{.\xspace}
\newcommand{\eg}{\emph{e.g}\OneDot}
\newcommand{\ie}{\emph{i.e}\OneDot}

\newcommand{\vs}{\emph{vs}\OneDot}

\makeatletter
\let\oldthebibliography\thebibliography
\renewcommand{\thebibliography}[1]{%
  \oldthebibliography{#1}%
  \setlength{\parskip}{0pt}%
  \setlength{\itemsep}{0pt plus 0.3ex}%
}
\makeatother
\title{ATTENTION-AWARE TRANSFORMER-BASED AGGREGATION NETWORK FOR VIDEO PERIOCULAR RECOGNITION}
\name{Luiz G F Carreira$^{\star}$, Breno A Mariano$^{\star}$, Victor H C de Melo$^{\star}$, David Menotti$^{\dagger}$, William Robson Schwartz$^{\star}$}
\address{$^{\star}$Department of Computer Science, Federal University of Minas Gerais, Belo Horizonte, Brazil \\
$^{\dagger}$Department of Informatics, Federal University of Paraná, Curitiba, Brazil}
\begin{document}
%\ninept
%
\maketitle
\begin{abstract}
Video periocular recognition is the task of recognizing an individual's identity based on the region around an individual's eyes. The periocular area is one of the most discriminative regions of the human face, making it suitable for recognition tasks. Its use as a biometric modality has emerged as an alternative, especially in surveillance scenarios where conventional biometric traits such as face or iris recognition become unfeasible due to unconstrained acquisition conditions. This paper proposes an attention-aware approach for video-based periocular recognition in surveillance environments. The framework consists of two main modules: feature embedding and aggregation. The feature embedding module is a deep convolutional neural network that maps periocular data to feature vectors. The aggregation module is an encoder-only transformer that adaptively learns to aggregate frame-level features into a single video representation and a feature vector for the still reference image. Experiments on the publicly available COX Face dataset show the robustness of the proposed method, consistently outperforming naive aggregation schemes. In the best scenario, the approach achieves $99.8\%$ of TPR@$1e^{-1}$ and $96.6\%$ of Rank-5.
\end{abstract}
\section{Introduction}
\label{sec:intro}

Video periocular recognition has received much attention from the scientific community in recent years. This research area has become a relevant study field, mainly because it plays an important role in many real-world applications such as visual surveillance~\cite{luz2018deep, kim2018convolutional} and video search.

Compared to single still image-based periocular recognition~\cite{miller2010performance, park2009periocular, zhao2016accurate, talreja2022attribute, zanlorensi2022new}, video-based periocular recognition can exploit additional useful information across different frames of the same subject. However, frames from the same individual can exhibit significant variations, mainly in surveillance scenarios, due to uncontrolled and non-cooperative frame acquisition processes, such as blur, occlusion, varied illuminations, and pose changes, which can create a mismatch between the target and probe data. Hence, the key issue in video-based periocular recognition is to learn a video-level representation that effectively combines information from different frames, discarding noisy information while maintaining discriminative features.

In recent years, many research efforts have been dedicated to the development of methods capable of integrating information across different images to obtain a video-level representation~\cite{liu2019feature, taigman2014deepface}, whether using different pooling strategies, such as average pooling and max pooling, or focusing on high-quality biometric data while discarding low-quality ones~\cite{qi2018cnn, qi2018boosting}, or that adaptively consider different levels of importance to frames. In general, most approaches have focused either on facial recognition or on naive pooling strategies, and have been applied to controlled scenarios with very little variability in the biometric quality of the data, while ignoring periocular recognition. Therefore, video periocular recognition is still an open research area and remains unexplored. Our intuition is that an ideal framework must highlight valuable biometric data while suppressing noisy components.

\begin{figure*}[t]
    \centering
    \includegraphics[width=0.85\textwidth]{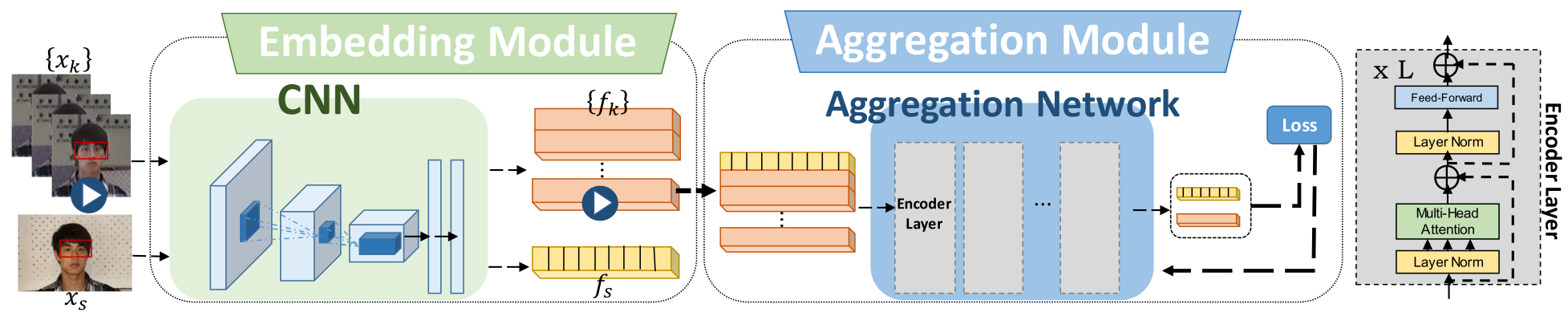}
    \caption{Overview of the proposed framework. The embedding module extracts frame-level and still image feature vectors, which are subsequently concatenated and then fed into the aggregation module to produce a video-level and a single still image representations.}
    \label{fig:architecture}
\end{figure*}

Accordingly~\cite{huang2015coxdataset}, there are three distinct video-based recognition tasks: video-to-still (V2S), still-to-video (S2V), and video-to-video (V2V). In the V2S scenario, the algorithm matches a query video against a still target image. In this paper, we propose a novel attention-based aggregation framework for video periocular recognition in surveillance environments between different modalities, \ie, V2S. Our method adaptively learns a new representation for the still target image as well as a video-level representation based on the relationships between the video frames and the still image.

The main contributions of this paper are: i) A novel feature aggregation method for video periocular recognition capable of handling a variable number of video frames and producing fixed-size representations that are invariant to the image order. To the best of our knowledge, this is one of the first works to investigate an attention-based feature aggregation scheme for video surveillance using the periocular region and ii) The introduction of a feature aggregation framework composed of two main components: an embedding and an aggregation module, which can be trained either end-to-end or separately.
%These guidelines include complete descriptions of the fonts, spacing, and
%related information for producing your proceedings manuscripts. Please follow
%them and if you have any questions, direct them
%to \\\texttt{papers@2026.ieeeicip.org}.

\section{RELATED WORKS}
\label{sec:relatedworks}

Since the periocular region is extracted from facial images, both video-based and image set-based face and periocular recognition share several common characteristics. In this work, we consider recognition methods based on video or image sets that exploit periocular or facial data.

Regarding image set-based recognition studies, previous works have proposed representing a face image set as manifolds in the feature space and computing similarity or dissimilarity metrics to perform recognition~\cite{lee2003video, arandjelovic2005face, kim2007boosted, wang2008manifold}. These traditional methods have shown interesting results under well-controlled conditions. However, in uncontrolled scenarios, such approaches cannot handle facial appearance variations, which significantly reduce their performance.

To use information from all video frames without discarding low biometric-quality ones, some simple aggregation strategies have been adopted in video face or periocular recognition. The authors in~\cite{taigman2014deepface} propose to sample 100 random pairs of frames for each test pair, one from each video, and use the mean of the weighted $\chi^2$ similarity scores. In~\cite{schroff2015facenet}, the authors compute the average similarity over all pairs formed from the first one hundred detected face frames in each video. On the other hand, instead of computing pairwise frame feature similarities and then fusing the matching results (\eg,~\cite{taigman2014deepface, schroff2015facenet}), other methods have proposed max or average pooling to aggregate frame-level features. DPR-V2S~\cite{carreira2024dpr} is one of the first approaches to apply average pooling in video periocular recognition, and thus it is used as a baseline for our work. However, although this method has shown better results than approaches that do not use information from all frames, we believe that average pooling is unable to effectively handle noisy features from low biometric-quality frames. This observation motivates us to propose a new approach to video periocular recognition.

The main challenge of naive aggregation methods (\eg, max or average pooling) is that they assign equal relevance to all video frames, regardless of their quality. To mitigate the impact of low biometric-quality frames during the aggregation step, more robust aggregation strategies have been proposed. NAN~\cite{yang2017neural} introduces an attention mechanism to adaptively assign weights to frames, thereby reducing the contribution of low-quality frames in the final video-level representation. However, NAN assumes that all elements of a given feature vector have equal importance. Inspired by NAN, the work by Liu et al~\cite{liu2019feature} proposes a generalized aggregation scheme that uses a modified attention mechanism to adaptively weight each feature separately.

Similar to NAN~\cite{yang2017neural} and~\cite{liu2019feature}, we present an aggregation strategy to adaptively assign weights to frames with the flexibility to focus more on some features than others. However, unlike~\cite{yang2017neural} and~\cite{liu2019feature}, our method leverages the self-attention mechanism from the encoder-only transformer model.

\section{Proposed Approach}
\label{sec:method}

Our aggregation framework is composed of two main modules: the feature embedding and the aggregation (Figure~\ref{fig:architecture}). The first component is a deep convolutional network that acts as a feature extractor. The second part is the aggregation module, which combines the feature vectors of all video frames (query) to form a video-level feature vector and learns a new representation for the still target image based on their relationships, \ie, query and target data. Thus, the framework becomes more robust for video periocular recognition.

\subsection{Feature Embedding Module}
\label{ssec:featureembedmodule}

The embedding module is built upon a deep CNN model. We adopt the pretrained ResNet-50 model on the VGGFace2 data set to embed each frame of the video and the still image in a representation of periocular features. The ResNet-50 produces 512-dimension periocular feature vectors, which are concatenated and then forwarded to the aggregation module.

\subsection{Aggregation Module}
\label{ssec:aggregationmodule}

Assume the video periocular recognition task on n pairs of periocular data, \ie, video data $(X^i,y_i)_{i=1}^n$ and a still image $(x_s^i,y_i)$, where $X^i$ is a video sequence or a image set and $K_i$ represents a varying image number with $X^i=\{x_1^i,x_2^i,...,x_{k_i}^i\}$ in which $x_k^i, k=1,...,K_i$ is the $k$-th frame in the video, $x_s^i$ is the reference still image, and $y_i$ is the corresponding subject ID of both $X^i $ and $x_s^i$. Our goal is to generate a video-level and a single still image feature vector representation by using all the frames of the video and the still image based on their relationships.

\vspace{1mm}
\noindent \textbf{The proposed aggregation strategy:} The input data to our encoder-only transformer-based aggregation network is a set of vectors $\{f_k\}$ of dimensionality $D$, where $k=1,...,K_{i+1}$. Sometimes, we refer to these data vectors as \textit{tokens}. Each frame $x_k^i$ and the still image $x_s^i$ have a corresponding convolutional feature representation $f_k^i$ and $f_s^i$ extracted from the feature embedding module respectively. The first step is to concatenate these feature vectors to form a matrix of genuine samples $F^i$, \ie,

\begin{equation}
\mathbf{F}^{i}
=
\begin{bmatrix}
\mathbf{f}_s^{i^\top} \\
\mathbf{f}_1^{i^\top} \\
\vdots \\
\mathbf{f}_{k_i}^{i^\top} \\
\end{bmatrix}_{K_{i+1} \times D}
=
\begin{bmatrix}
\mathbf{f}_1^{i^\top} \\
\mathbf{f}_2^{i^\top} \\
\vdots \\
\mathbf{f}_{k_{i+1}}^{i^\top} \\
\end{bmatrix}
\end{equation}

For better readability, we substitute $f_s^i$ with $f_1^i$, $f_1^i$ with $f_2^i$ and so on. Suppose that we replace $f_s^i$ with $f_s^j$, where $i \ne j$, then we have a matrix of impostor sample $F^i$. Now, we have a set of input tokens $f_s^i,f_1^i, ...,f_{k_i}^i$ in an embedding space. We want to map this to a new space $y_s^i,y_1^i,...,y_{k_i}^i$ that captures a richer semantic structure while keeping the same number of tokens. The value of each $y_k^i$ depends not just on the corresponding input vector $f_k^i$ but on all the input tokens in the set. We address this using the self-attention mechanism. Each output $y_k^i$ is a linear combination of the input vectors.

\begin{equation}
\mathbf{y_k^i} = \sum_{m=1}^{K_{i+1}} a_{km}^i \mathbf{f}_m^i
\label{eq:linearcombinationattention}
\end{equation}

\begin{equation}
a_{km}^i =
\frac{\exp\!\left(\mathbf{f}_k^{i^\top}\mathbf{f}_m^i\right)}
{\sum_{m'=1}^{K_{i+1}} \exp\!\left(\mathbf{f}_k^{i^\top}\mathbf{f}_{m'}^i\right)}
\end{equation}

where $a_{km}^i$ is the attention coefficient of the $k^i$-th token related to the $m^i$-th input data. If the $a_{km}^i$ weight has a value close to zero, then the $f_m^i$ token has little influence on the output vector $y_k^i$ , whereas a large value indicates high impact.

We can write Equation~\eqref{eq:linearcombinationattention} in matrix notation by using the data matrix $F^i$, $Q^i$, $K^i$, and $V^i$

\begin{equation}
\begin{aligned}
\mathbf{Q^i} &= \mathbf{F^i}\mathbf{W}^{(q)},\,\,
\mathbf{K^i} &= \mathbf{F^i}\mathbf{W}^{(k)},\,\,
\mathbf{V^i} &= \mathbf{F^i}\mathbf{W}^{(v)}
\end{aligned}
\label{eq:qkvmatrices}
\end{equation}

\vspace{-2em}

\begin{equation}
\mathbf{Y^i}
=
\mathrm{Attn}(\mathbf{Q^i}, \mathbf{K^i}, \mathbf{V^i})
=
\mathrm{Softmax}\!\left(
\frac{\mathbf{Q^i}\mathbf{K}^{i^\top}}{\sqrt{D_k}}
\right)\mathbf{V^i}.
\label{eq:matrixattention}
\end{equation}

Note that to determine the query $Q^i$, key $K^i$, and value $V^i$ matrices, we use the same input sequence $F^i$. This process is self-attention. Here, the matrix $W^k$ has dimensionality $D \; X \; D_k$ where $D_k$ is the length of the key vector. A common choice is $D \; = \; D_k$.

In practice, the encoder-only transformer architectures have multiple layers, and in each layer there are $H$ heads indexed by $h \: = 1,...,H$. Therefore, we can formulate the attention in each head as $\mathbf{H_h^i} = \mathrm{Attn}(\mathbf{Q_h^i}, \mathbf{K_h^i}, \mathbf{V_h^i})$, where $\mathrm{Attn(.,.,)}$ is given by \eqref{eq:matrixattention}, so that we can rewrite \eqref{eq:qkvmatrices} and \eqref{eq:matrixattention}

\begin{equation}
\begin{aligned}
\mathbf{Q_h^i} &= \mathbf{F^i}\mathbf{W}_h^{(q)},\,
\mathbf{K_h^i} &= \mathbf{F^i}\mathbf{W}_h^{(k)},\,
\mathbf{V_h^i} &= \mathbf{F^i}\mathbf{W}_h^{(v)}
\end{aligned}
\label{eq:qkvmatricesmodified}
\end{equation}

\vspace{-1.2em}

\begin{equation}
\mathbf{Y^i(F^i)}
=
\mathrm{Concat}[\mathbf{H_1^i}, ..., \mathbf{H_H^i}]\mathbf{W^o}.
\label{eq:matrixattentionmodified}
\end{equation}

The self-attention component forms the core of transformer models. However, other relevant parts include residual connections, pre-norm operations, and nonlinear neural network (MLP). Thus, the model output can be formulated as $\mathbf{Z}^i = \mathbf{Y^i}(\mathbf{F^i}') + \mathbf{F^i},\quad \text{where } \mathbf{F^i}' = \mathrm{LayerNorm}[\mathbf{F^i}]$ and $\tilde{\mathbf{F}^i} = \mathrm{MLP}(\mathbf{Z^i}') + \mathbf{Z^i},\quad \text{where } \mathbf{Z^i}' = \mathrm{LayerNorm}[\mathbf{Z^i}]$. $\tilde{\mathbf{F}}^i$ denotes the set of output feature representations. Therefore, the final video-level and still image representations are, respectively, $\mathbf{r^i} = \frac{1}{K_{i}} \sum_{k=2}^{K_{i+1}} \tilde{\mathbf{f}_k^i}$, and $\mathbf{r_s^i} = \tilde{\mathbf{f}_1^i}$.

\begin{table*}[t]
\centering
\caption{Average performance evaluation on the COX Face dataset over 10 splits. For verification, the true acceptance rates (TAR) \vs false acceptance rates (FAR) are reported. For identification, the Rank-$N$ accuracies are presented. ($^{\dagger}$: our method with the frames presented in the reversed order.)}
\label{tab:cox}
\footnotesize
\setlength{\tabcolsep}{12pt}
\begin{tabular}{c l ccc c cc}
\hline
\multirow{2}{*}{Exp.} &\multirow{2}{*}{Method }
& \multicolumn{3}{c}{1:1 Verification TAR}
&
& \multicolumn{2}{c}{1:N Identification Rank-N}\\
\cline{3-5} \cline{7-8}
& 
& FAR=0.001 & FAR=0.01 & FAR=0.1
&
& Rank-1 & Rank-5 \\
\hline

\multirow{5}{*}{V1-S}
& DPR-V2S
& 0.752 $\pm$ 0.010 & 0.836 $\pm$ 0.007 & 0.947 $\pm$ 0.007 
&
& 0.722 $\pm$ 0.010 & 0.855 $\pm$ 0.007\\

& CNN+MaxPool
& 0.672 $\pm$ 0.048 & 0.835 $\pm$ 0.035 & 0.952 $\pm$ 0.017
&
& 0.684 $\pm$ 0.010 & 0.870 $\pm$ 0.008 \\

& CNN+Random
& 0.348 $\pm$ 0.017 & 0.521 $\pm$ 0.020 & 0.767 $\pm$ 0.023
&
& 0.383 $\pm$ 0.010 & 0.550 $\pm$ 0.012 \\

& CNN+Pairwise
& 0.417 $\pm$ 0.020 & 0.686 $\pm$ 0.026 & 0.930 $\pm$ 0.009
&
& 0.568 $\pm$ 0.009 & 0.790 $\pm$ 0.011 \\

& Ours$^{\dagger}$
& 0.766 $\pm$ 0.037 & 0.929 $\pm$ 0.021 & 0.994 $\pm$ 0.003
&
& 0.797 $\pm$ 0.009 & 0.941 $\pm$ 0.006 \\

& Ours
& \textbf{0.772} $\pm$ 0.037 & \textbf{0.929} $\pm$ 0.020 & \textbf{0.994} $\pm$ 0.003
&
& \textbf{0.799} $\pm$ 0.009 & \textbf{0.942} $\pm$ 0.006 \\

\hline

\multirow{5}{*}{V2-S}
& DPR-V2S
& \textbf{0.671} $\pm$ 0.009 & 0.757 $\pm$ 0.010 & 0.922 $\pm$ 0.008 
&
& 0.658 $\pm$ 0.013 & 0.795 $\pm$ 0.008 \\

& CNN+MaxPool
& 0.599 $\pm$ 0.046 & 0.779 $\pm$ 0.022 & 0.926 $\pm$ 0.013
&
& 0.645 $\pm$ 0.014 & 0.798 $\pm$ 0.009 \\

& CNN+Random
& 0.342 $\pm$ 0.044 & 0.582 $\pm$ 0.032 & 0.816 $\pm$ 0.018
&
& 0.403 $\pm$ 0.019 & 0.580 $\pm$ 0.019 \\

& CNN+Pairwise
& 0.495 $\pm$ 0.030 & 0.696 $\pm$ 0.028 & 0.910 $\pm$ 0.012
&
& 0.576 $\pm$ 0.011 & 0.744 $\pm$ 0.012 \\

& Ours$^{\dagger}$
& 0.627 $\pm$ 0.042 & 0.870 $\pm$ 0.023 & 0.984 $\pm$ 0.007
&
& 0.752 $\pm$ 0.013 & 0.896 $\pm$ 0.008 \\

& Ours
& 0.628 $\pm$ 0.044 & \textbf{0.874} $\pm$ 0.024 & \textbf{0.984} $\pm$ 0.008
&
& \textbf{0.752} $\pm$ 0.013 & \textbf{0.896} $\pm$ 0.007 \\

\hline

\multirow{5}{*}{V3-S}
& DPR-V2S
& 0.830 $\pm$ 0.006 & 0.894 $\pm$ 0.008 & 0.980 $\pm$ 0.000
&
& 0.746 $\pm$ 0.001 & 0.878 $\pm$ 0.005 \\

& CNN+MaxPool
& 0.740 $\pm$ 0.044 & 0.883 $\pm$ 0.023 & 0.984 $\pm$ 0.006
&
& 0.748 $\pm$ 0.011 & 0.901 $\pm$ 0.004 \\

& CNN+Random
& 0.433 $\pm$ 0.050 & 0.697 $\pm$ 0.044 & 0.914 $\pm$ 0.018
&
& 0.499 $\pm$ 0.014 & 0.703 $\pm$ 0.009 \\

& CNN+Pairwise
& 0.592 $\pm$ 0.060 & 0.877 $\pm$ 0.025 & 0.975 $\pm$ 0.010
&
& 0.721 $\pm$ 0.010 & 0.869 $\pm$ 0.005 \\

& Ours$^{\dagger}$
& 0.862 $\pm$ 0.020 & 0.971 $\pm$ 0.007 & 0.998 $\pm$ 0.001
&
& 0.872 $\pm$ 0.007 & 0.966 $\pm$ 0.005 \\

& Ours
& \textbf{0.864} $\pm$ 0.020 & \textbf{0.971} $\pm$ 0.007 & \textbf{0.998} $\pm$ 0.001
&
& \textbf{0.872} $\pm$ 0.006 & \textbf{0.966} $\pm$ 0.005 \\

\hline
\end{tabular}
\end{table*}

\subsection{Network Training}
\label{ssec:networktraining}

\textbf{Training loss:} Our aggregation network can be used either for video periocular verification and identification. For verification, we trained our aggregation network based on the deep metric learning paradigm. Once learned, instead of trying to output a single label, the model outputs two real-valued feature vectors, a final video-level and still image deep convolutional feature vectors. In simple words, the model is trying to learn a new space where representations of the same individual are close to each other in this new space, and samples of different subjects are farther apart. We use the cosine embedding loss: $\mathrm{loss}(x,y)=1-\cos(x_1,x_2)\;\text{if }y=1,\; \max(0,\cos(x_1,x_2)-\text{margin})\;\text{if }y=-1$, where $y=1$ if the pair $x_1,x_2$ is from the same identity and $y=-1$ otherwise. The constant $\mathrm{margin}$ is set to $0.5$ in all experiments. This constant indicates how much our algorithm tolerates of the cosine similarity value between impostor pairs.

The identification task requires training a multi-class classifier to assign a particular identity to the given input data among all identities in the database. However, this type of identification system is not scalable if the number of individuals changes much in the database. Thus, a more efficient and effective way of performing identification is via verification.  Given a query data, we run our verification algorithm one time for each reference image in the gallery. 

\noindent \textbf{Module training:} The feature embedding and aggregation modules can be trained jointly or in separate stages. In this work, we follow the latter strategy. First, the CNN is trained on single images as a multi-class classifier, and then it is used as a periocular feature extractor. Afterwards, the aggregation network is trained on the periocular features produced by CNN under a deep metric learning framework. We adopted this separate strategy to focus on analyzing the robustness and performance of the aggregation network with the self-attention mechanism for video periocular recognition. We leave the end-to-end training strategy as our future work.

\section{Experimental Results}
\label{sec:exp}

\textbf{Datasets and protocols:} The VGGFace2~\cite{cao2018vggface2} dataset which includes around 3.31M images of about 92k classes is used to train our feature embedding module for the still image-based periocular recognition task. To train and evaluate our aggregation module, we use the COX Face~\cite{huang2015coxdataset} database, which contains 1k subjects and 3k videos. We adopt the same evaluation protocol for the V2S scenario defined by the authors~\cite{huang2015coxdataset}. In this protocol, we have 10 random partitions where each fold has 300/700 (train/test) individuals. All the following results reflect the average of the 10 trials.

\vspace{0.5mm}
\noindent  \textbf{Embedding module training:} As mentioned in Section~\ref{ssec:featureembedmodule}, for the embedding network, we employ ResNet50 pre-trained on the VGGFace2 dataset. We remove the classifier layers and add the BN-Dropout-Fc-BN structure before the last layer (Softmax) to get the final 512-D feature vector. We set the dropout to $0.5$ and do not freeze the layers of the base model, allowing the entire model to adapt to the new domain. MTCNN is used to detect and align the face images. After that, we follow~\cite{kim2018convolutional} to generate a single periocular region crop from a single face image based on the eye's distance, \ie, we create periocular region data that encompass both left and right eyes. Before rescaling the crop of the periocular region to the size $224 \ X \ 224$, we add zero padding layer at the top and bottom sides of the image/frame to get the $1:1$ aspect ratio, so that, we avoid the creation of artifacts, \ie, undesirable alterations that appear in digital images during the rescaling.

\begin{figure*}[t]
    \centering

    % Linha 1
    \subfloat[]{\includegraphics[width=0.3\textwidth,trim=1 1 1 1,clip]{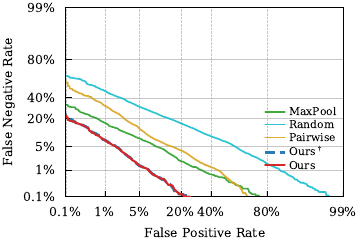}}%\hfill
    \subfloat[]{\includegraphics[width=0.3\textwidth,trim=1 1 1 1,clip]{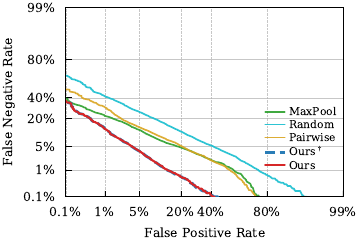}}%\hfill
    \subfloat[]{\includegraphics[width=0.3\textwidth,trim=1 1 1 1,clip]{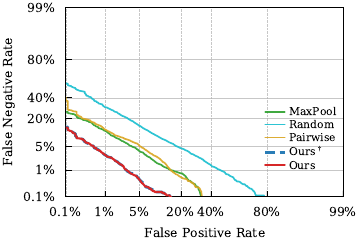}}

    % Linha 2
    \subfloat[]{\includegraphics[width=0.3\textwidth,trim=7 7 7 7,clip]{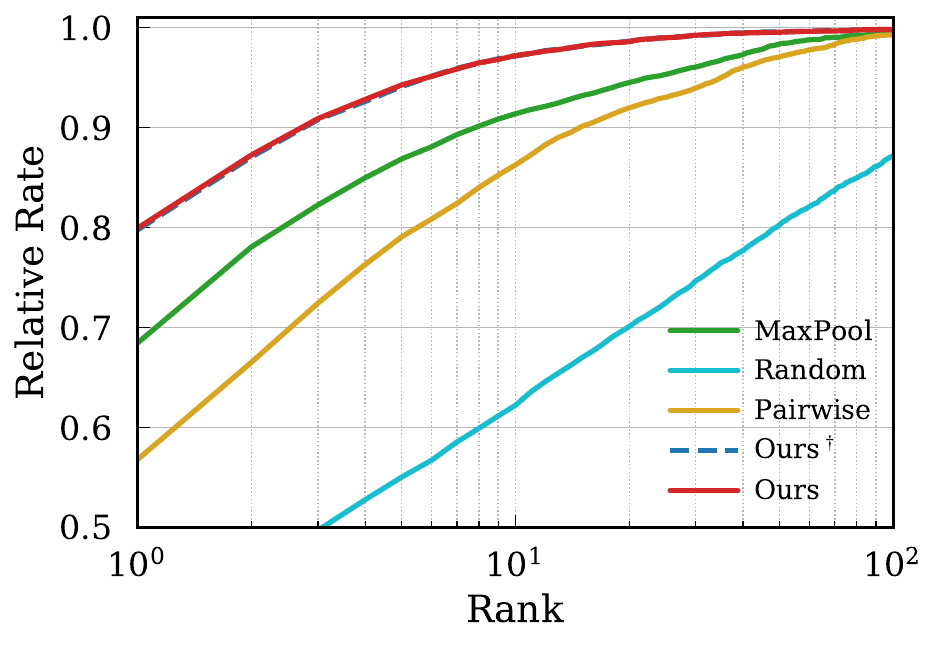}}%\hfill
    \subfloat[]{\includegraphics[width=0.3\textwidth,trim=7 7 7 7,clip]{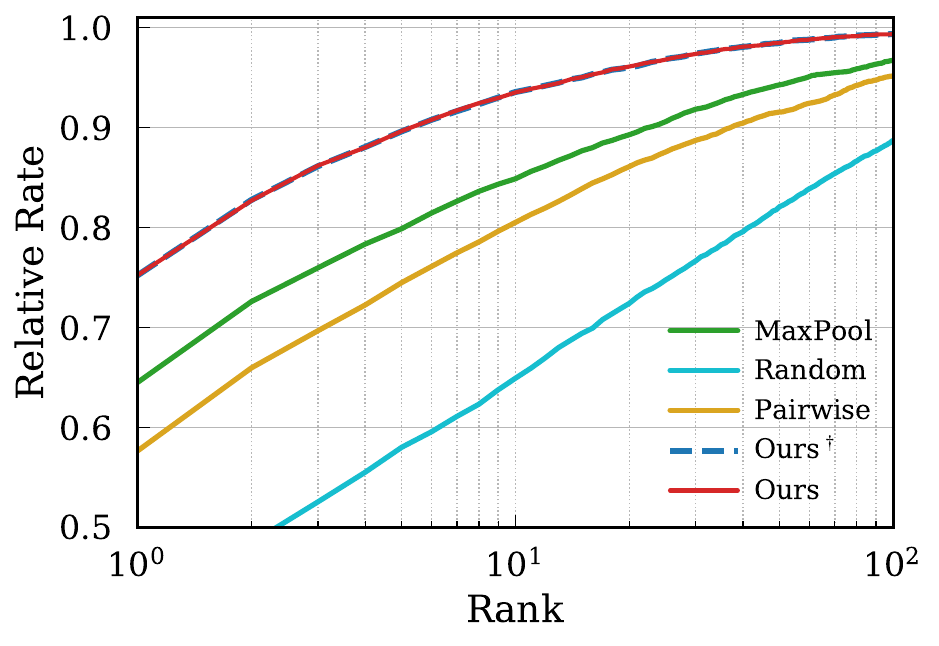}}%\hfill
    \subfloat[]{\includegraphics[width=0.3\textwidth,trim=7 7 7 7,clip]{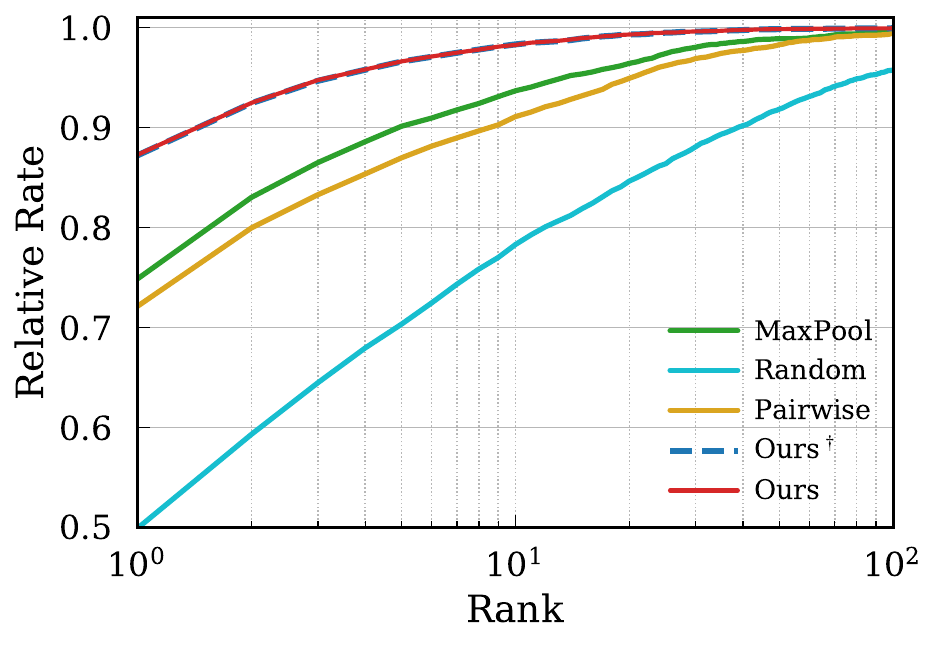}}
    \vspace{-1.em}
    \caption{Average DET (Top) and CMC (Bottom) curves of our method and the baselines on the COX Face dataset for the three cameras (\ie first column represents the V1-S and so on) over 10 splits.}
    \label{fig:6plots}
\end{figure*}

\vspace{0.5mm}
\noindent \textbf{Aggregation module training:} To train the encoder-only transformer-based aggregation network, we use the COX Face dataset. Our transformer model has 16 heads in each encoder layer and has 4 encoder layers. The model is learned with AdamW optimizer with weight decay $1e^{-1}$, batch size of 16, and learning rate of $1e{-5}$. We train the network during 10 epochs and create 1/15 (genuine/impostor) pairs for each subject in the training database.

\vspace{0.5mm}
\noindent  \textbf{Baseline methods:} As baseline methods, we consider max pooling, as well as random and pairwise comparison strategies. In the random strategy, a single frame is randomly selected and compared to the still target, whereas in the pairwise strategy, each video frame is individually compared to the still target, and the resulting scores are averaged. Cosine distance is used to compute the dissimilarity score between the query data (\ie, the video-level representation, individual frames, or a randomly selected frame) and the still reference image. For fairness, all baseline methods and our proposed approach employ the same feature embedding module. Additionally, we include DPR-V2S~\cite{carreira2024dpr} as a baseline.

\begin{figure}[htb]
\begin{minipage}[b]{1.0\linewidth}
  \centering
  \centerline{\includegraphics[width=0.75\textwidth]{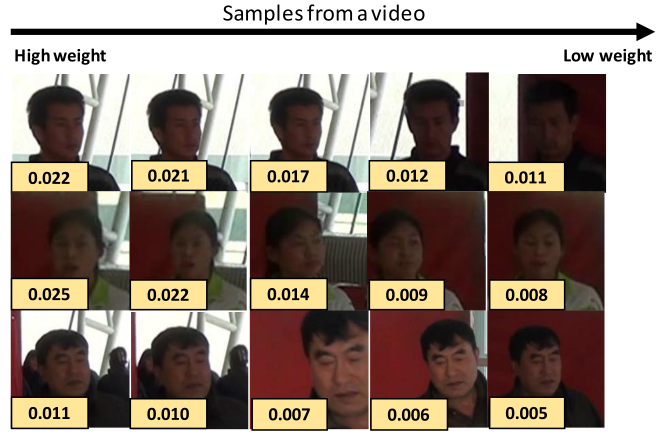}}
\end{minipage}
\vspace{-2.em}
\caption{Samples from the Cox Face dataset showing the weights of the video frames computed by our aggregation network. Each row shows five frames sorted by their weights.}
\label{fig:qua}
\end{figure}

\subsection{Results on the COX Face Dataset}
\label{ssec:resultscox}

We test the proposed method on both verification and identification protocols. For verification, we report the true acceptance rate (TAR) \textit{vs.} false positive rate (FAR). For identification, the Rank-N accuracies are reported. Table~\ref{tab:cox} presents the results of different methods for all three experiments (V1-S, V2-S, and V3-S) and Figure~\ref{fig:6plots} shows the Detection Error Tradeoff (DET) curves for verification as well as the cumulative match characteristic (CMC) curves for identification.

In general, \textit{CNN+Random} performs worst among all the baseline methods. For all three experiments and all metrics, our approach consistently outperforms the baseline strategies, \ie, \textit{CNN+MaxPool}, \textit{CNN+Random}, and \textit{CNN+Pairwise}. For verification, our method surpasses DPR-V2S by about $11\%$, $15\%$, and $9\%$ for the V1-S, V2-S, and V3-S experiments to the TPR@$1e^{-2}$ metric respectively. For identification evaluation, our approach outperforms the DPR-V2S by about $11\%$, $14\%$, and $17\%$ for the Rank-1 metric, and by about $10\%$, $13\%$, and $10\%$ for the Rank-5 metric, for V1-S, V2-S, and
V3-S experiments respectively.

Figure~\ref{fig:qua} exhibits some typical examples of the weighting results extracted from the attention head. Each weight reflects the importance that the still reference image of a given subject assigns to the video frames. Our aggregation network has the ability to focus on frames in which the subject’s eyes are open and the gaze is near-frontal, avoiding downward or upward gaze, which would compromise recognition.

\section{Conclusions}
\label{sec:conclusion}

This paper presented a novel feature aggregation scheme for video periocular recognition that consistently outperforms all baseline methods in the COX Face dataset. The experimental results show statistically significant gains under both verification and identification protocols, demonstrating the effectiveness and robustness of the proposed approach. Our method is designed to suppress irrelevant frame-level information while preserving discriminative features. Thus, we showed that employing an attention mechanism to leverage and integrate information from both low- and high-quality frames leads to improved overall periocular recognition performance.

\section*{Acknowledgments}
The authors would like to thank the National Council for Scientific and Technological Development -- CNPq (Grant 312565/2023-2). This study was financed in part by the Coordenação de Aperfeiçoamento de Pessoal de Nível Superior - Brasil (CAPES) - Finance Code 001. The work of David Menotti was supported by the National Council for Scientific and Technological Development (CNPq) under Grant 315409/2023-1.

\bibliographystyle{IEEEbib}
\bibliography{strings,refs}

\end{document}